\documentclass[lettersize,journal]{IEEEtran}

\usepackage{amsmath,amsfonts}
\usepackage{algorithm}
\usepackage{algorithmic}
\usepackage{array}
\usepackage{textcomp}
\usepackage{stfloats}
\usepackage{url}
\usepackage{verbatim}
\usepackage{graphicx}
\usepackage{cite}
\usepackage{color}
\usepackage{xcolor}
\usepackage{ragged2e}
\usepackage{threeparttable}
\usepackage{float}
\usepackage{placeins}
\usepackage{booktabs}
\usepackage{multirow}
\usepackage[hidelinks]{hyperref}

\usepackage[caption=false,font=footnotesize]{subfig}

\begin{document}

\title{A Lightweight Multi-Scale Attention Framework for Real-Time Spinal Endoscopic Instance Segmentation}

\author{Qi Lai, 
         JunYan Li ,
         Qiang Cai ,
         Lei Wang,  ~\IEEEmembership{Senior Member, IEEE,} 
         Tao Yan, \\
         and XiaoKun Liang , \IEEEmembership{Member, IEEE}

\thanks{Manuscript received Dec 23, 2025;}}



\maketitle

\begin{abstract}
Real-time instance segmentation for spinal endoscopy is important for identifying and protecting critical anatomy during surgery, but it is difficult because of the narrow field of view, specular highlights, smoke/bleeding, unclear boundaries, and large scale changes. Deployment is also constrained by limited surgical hardware, so the model must balance accuracy and speed and remain stable under small-batch (even batch-1) training.
We propose LMSF-A, a lightweight multi-scale attention framework co-designed across backbone, neck, and head. The backbone uses a C2f-Pro module that combines RepViT-style re-parameterized convolution (RVB) with efficient multi-scale attention (EMA), enabling multi-branch training while collapsing into a single fast path for inference. The neck improves cross-scale consistency and boundary detail using Scale-Sequence Feature Fusion (SSFF) and Triple Feature Encoding (TFE), which strengthens high-resolution features. The head adopts a Lightweight Multi-task Shared Head (LMSH) with shared convolutions and GroupNorm to reduce parameters and support batch-1 stability.
We also release the clinically reviewed PELD dataset (61 patients, 610 images) with instance masks for adipose tissue, bone, ligamentum flavum, and nerve. Experiments show that LMSF-A is highly competitive (or even better than) in all evaluation metrics and much lighter than most instance segmentation methods requiring only 1.8M parameters and 8.8 GFLOPs, and it generalizes well to a public teeth benchmark.
Code and dataset: https://github.com/hhwmortal/PELD-Instance-segmentation.
\end{abstract}


\section{Introduction}
\label{sec:intro}

Spinal endoscopy has become a cornerstone of minimally invasive spine surgery due to reduced tissue trauma and faster postoperative recovery \cite{choi2017endoscopic,zhan2025comparative, cleary2002technology}. During these procedures, \textit{real-time and accurate delineation} of key anatomical structures—such as the ligamentum flavum, nerves, bone, and adipose tissue—is critical for avoiding iatrogenic injury and guiding safe dissection \cite{simpson2022spinal}.

\textit{Instance segmentation} provides pixel-level localization with explicit separation of individual structures, enabling actionable overlays for intraoperative navigation and decision support \cite{he2017mask,bolya2019yolact,wang2020solov2,liu2021swin,cheng2022masked,woo2023convnext,varghese2024yolov8}. 
However, spinal endoscopic images constitute a particularly challenging visual domain, characterized by a narrow field of view (FoV), non-uniform illumination and specular highlights, smoke and bleeding artifacts, strong inter-class similarity in color and texture, ambiguous boundaries, and substantial scale variation \cite{hussain2022challenges}. 
Achieving robust, real-time instance segmentation under these conditions—especially on resource-constrained surgical hardware—remains an open problem \cite{burkett2024advances}.

Deep learning has driven major progress in medical image segmentation, evolving from semantic to instance-level understanding \cite{li2024spatial,ming2025few,yao2025prompting}. Two-stage detectors (e.g., Mask R-CNN \cite{he2017mask}) often deliver strong accuracy, but their parameter count and inference latency limit real-time use \cite{sun2023sparse}. In contrast, single-stage methods (e.g., YOLACT \cite{bolya2019yolact}, SOLOv2 \cite{wang2020solov2}, and YOLO variants \cite{varghese2024yolov8, yolo11_ultralytics}) are typically faster and easier to deploy, yet frequently struggle with boundary precision and sensitivity to small, clinically critical structures in complex endoscopic backgrounds. Vision Transformers \cite{liu2023survey,han2021transformer} and hybrid architectures enhance global context modeling, but their computational and memory demands complicate stable deployment on edge devices such as endoscopy workstations or embedded platforms. Although many lightweight strategies have been explored—via re-parameterized convolutions, streamlined attention, and modified neck/head designs—there is still a lack of a coherent end-to-end solution that simultaneously improves fine-grained boundary fidelity, ensures consistent multi-scale fusion, and remains robust under small-batch training and deployment in spinal endoscopy.

We summarize three key limitations of existing approaches:
\begin{enumerate}
\item \textbf{Accuracy--efficiency trade-off under surgical constraints:} Two-stage and Transformer models are often too heavy for real-time endoscopy; lightweight CNNs run fast but degrade on boundaries and small, clinically critical structures (e.g., thin ligament edges, nerve margins).
\item \textbf{Insufficient multi-scale and fine-grained fusion:} Standard FPN/PAN can misalign semantics and details across scales, causing oversmoothing, mask leakage, or missed small instances under noise, glare, or high tissue similarity.
\item \textbf{Fragility under small-batch and domain variability:} BatchNorm suffers at small batch sizes common in medical imaging; class imbalance and domain variability (anatomy, optics, illumination) hurt generalization; many attention modules overfocus on one dimension or require costly global interactions, reducing stability and deployability.
\end{enumerate}

To address these challenges, we propose \textbf{L}ightweight \textbf{M}ulti-\textbf{S}cale \textbf{F}usion with \textbf{A}ttention (LMSF-A), an end-to-end, lightweight architecture for spinal endoscopic instance segmentation that co-designs the backbone, neck, and head. The design is top–down: we first ease the accuracy–efficiency bottleneck in the backbone, then strengthen multi-scale fusion in the neck, and finally improve small-batch stability in the head for reliable deployment.

Specifically, resolve the accuracy–efficiency trade-off in feature extraction, we introduce C2f-Pro by embedding a Vision-Transformer–inspired re-parameterized block (RVB~\cite{wang2024repvit}) and efficient multi-scale attention into the C2f module. During training, multi-branch paths enrich representation; at inference, structural re-parameterization merges them into a single low-latency path. This preserves complementary local–global context without the heavy cost of full ViTs \cite{han2022survey}.

On this compact backbone, we enhance multi-scale and fine-grained fusion to recover small structures and sharpen boundaries. We add \textit{Scale-Sequence Feature Fusion} (SSFF) and \textit{Triple Feature Encoding} (TFE): SSFF forms a Gaussian scale sequence and uses 3D convolutions to learn cross-scale coherence, improving semantic–detail consistency; TFE asymmetrically aligns large/medium/small features via pooling-based downsampling and nearest-neighbor upsampling to keep high-frequency details. With selective RVB+EMA along the P3 path, the neck boosts small-object recall and boundary quality under noise, glare, and high tissue similarity.

In summary, the \textbf{contributions} of this paper are as follows:
\begin{itemize}
\item We proposed LMSF-A, a novel lightweight framework that achieves SOTA accuracy-efficiency trade-offs for real-time spinal endoscopic instance segmentation, requiring only 1.8M parameters and 8.8 GFLOPs while achieving 74.4\% mAP@50 and 189.3 FPS.

\item The newly designed modules comprise: C2f‑Pro, which integrates RepViT‑style re‑parameterized RVB and EMA to couple training‑time multi‑branch capacity with single‑path, low‑latency inference; a neck that combines SSFF (single‑pass, bidirectional, sequence‑aware cross‑scale fusion) and TFE (stride‑8 channel–spatial gating with optional edge‑aligned supervision) to refine boundaries and small structures; and LMSH, which employs a shared depthwise‑separable head with GroupNorm to improve parameter efficiency and batch‑1 stability.

\item We curate a clinically reviewed PELD instance segmentation dataset (610 images from 61 patients) with pixel-level masks for adipose tissue, bone, ligamentum flavum, and nerve, enabling standardized evaluation \footnote{Data available at \url{https://drive.google.com/file/d/1UXsnQyDAivstFrDTFr5t_YmmQGsx2m3P/view?usp=sharing}.}
\end{itemize}

\section{Related Works}
\label{sec:related}
\begin{figure*}[!htb]
\centering
\includegraphics[width=7.0in]{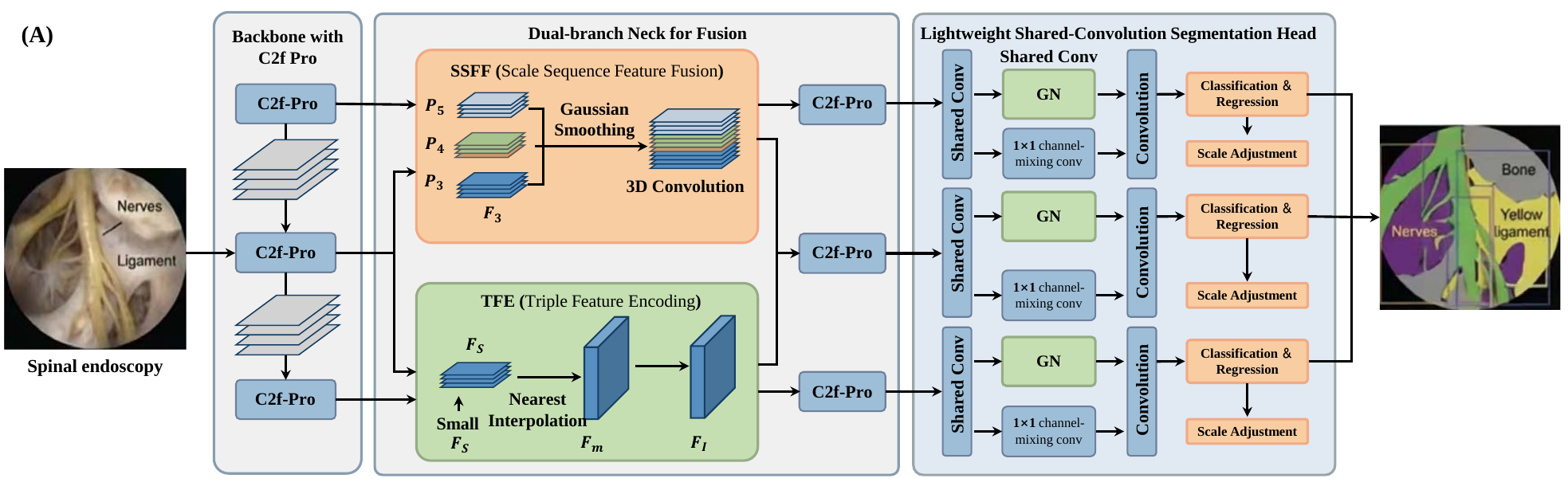}
\caption{Overall architecture of LMSF-A. GN denotes \textit{Group Normalization}.}
\label{fig:workflow}
\end{figure*}


\subsection{Instance Segmentation in Surgical and Endoscopic Scenes}
Instance segmentation for surgical and endoscopic images has evolved from proposal-based, two-stage pipelines to faster one-stage and hybrid designs that seek pixel-accurate masks with clear instance separation under dynamic intraoperative conditions \cite{simpson2022spinal,bolya2019yolact,wang2020solov2}. 
Two-stage models like Mask R-CNN \cite{he2017mask} remain strong for mask quality and small objects thanks to RoI-aligned features and multi-branch heads, but their latency and memory use often violate real-time OR constraints. One-stage methods—e.g., YOLACT \cite{bolya2019yolact} (prototypes with per-instance coefficients), SOLO/SOLOv2 \cite{wang2020solov2} (grid/set assignment), and Convnext \cite{woo2023convnext} (dynamic mask heads)—cut inference cost while keeping competitive accuracy, yet they can falter on thin or ambiguous boundaries, tissue adhesion, and large scale changes common in endoscopy \cite{hussain2022challenges}. 
Transformer \cite{han2021transformer} and hybrid models \cite{li2022hybrid} add global context but typically demand heavy compute and memory, limiting use on clinical edge devices. Enhancements such as stronger feature pyramids, boundary-aware losses, and video cues help, but reliable performance under glare, bleeding/irrigation, smoke, and rapid tool occlusions remains uncommon. These gaps call for architectures that preserve boundary fidelity and detect small structures while operating in real time.


\subsection{Lightweight Real-Time Vision for Dense Prediction}
Lightweight models are crucial for real-time inference under strict memory and power limits, especially in medical settings with small batch sizes and constrained hardware \cite{liu2024lightweight,shuvo2022efficient}. 
One effective strategy is structural re-parameterization: multi-branch blocks are used during training to boost representation capacity, then algebraically merged into a single-path convolution for fast, low-latency inference \cite{hu2022online,fei2024repan}. 
This is best paired with normalization schemes that remain stable at small batch sizes. Another approach is efficient attention, which balances local detail and broader context through channel-, spatial-, or scale-selective mechanisms while avoiding expensive global operations that slow models down \cite{khan2024esdmr}. 
A third direction is improved cross-scale fusion beyond standard FPN/PAN, using topology- or sequence-aware fusion and detail-preserving up/downsampling to reduce semantic–detail mismatch, improving small-instance recall and boundary sharpness \cite{liao2024lightweight}. 
Despite progress, few studies jointly evaluate these components for accuracy, latency, and stability under deployment-relevant conditions (e.g., batch size of 1 and surgical resolutions), or report full runtime profiles including end-to-end latency, memory footprint, and FP16/INT8 precision. These gaps motivate end-to-end designs that co-optimize the backbone, neck, and head for both segmentation quality and practical deployment.


\subsection{Spinal Endoscopy: Domain Challenges and Research Gaps}
Spinal endoscopy differs markedly from general laparoscopy \cite{simpson2022spinal}. 
The working corridor is narrow, and visibility is often degraded by fast-moving tools, strong glare, bleeding and irrigation fluids, bone dust, smoke, and high visual similarity among soft tissues such as dura, nerve, and epidural fat \cite{hussain2022challenges}. 
These factors make precise, real-time instance segmentation especially difficult \cite{bolya2019yolact}.
Early studies in this area focused on simpler tasks—instrument detection, tip localization, or single-class tissue segmentation. While they showed feasibility, they rarely addressed multi-class, instance-level delineation of clinically meaningful structures. 
Recent efforts have started to apply modern instance segmentation to percutaneous endoscopic lumbar discectomy (PELD), but comprehensive evaluations remain uncommon \cite{shuvo2022efficient}. In particular, few works: i) cover multiple anatomical classes, ii) benchmark two-stage, one-stage, and Transformer-based baselines under a unified protocol, and iii) analyze class-wise learnability and failure modes under realistic disturbances.
Practical deployment is also underexplored. Robustness with small-batch training and inference, handling of class imbalance, and the compute and memory limits of endoscopy workstations are often secondary concerns rather than core design constraints. These gaps call for a holistic, deployment-oriented study that combines clinically reviewed data and standardized metrics with architectures tailored for boundary quality, sensitivity to small structures, and real-time efficiency.


\section{Methodology}
\label{sec:methodology}

\subsection{LMSF-A Architecture Overview}
\label{ssec:arch_overview}

To improve both efficiency and practicality for spinal endoscopic instance segmentation, we propose LMSF-A, a lightweight attention-aware model composed of a compact backbone, a dual-branch neck, and a shared-convolution head (Figure~\ref{fig:workflow}). The pipeline is optimized for batch-1 inference, boundary fidelity, and small-structure sensitivity at surgical resolutions (640×640 by default).

\textbf{Backbone: Lightweight Multi-Scale Feature Extractor (LMFE) with C2f Pro.} We build on a streamlined CSPDarknet53 (light variant) as the main feature extractor to produce multi-scale features while balancing semantics and details. Each stage adopts the proposed C2f-Pro module, which integrates re-parameterized convolutions in a ViT-style block (RVB) together with a selective Efficient Multi-scale Attention (EMA) module at the deepest stage to enhance long-range context and robustness to specular highlights. Given an RGB image $\mathbf{I}\in\mathbb{R}^{3\times H\times W}$, the backbone yields a pyramid
\begin{equation}
\label{eq:Backbone}
(P_3,,P_4,,P_5)=\mathrm{Backbone}(\mathbf{I}),
\end{equation}
where $P_\ell\in\mathbb{R}^{C_\ell\times H_\ell\times W_\ell}$ ($\ell\in{3,4,5}$), and $P_3$, $P_4$, and $P_5$ are the feature maps at strides $s\in{8,16,32}$.

Given the backbone pyramid ${P_\ell}$ LMFE aligns channels to $C_f$ while preserving spatial resolution:
\begin{equation}
\label{eq:lmfe_align}
F_\ell=\phi(P_\ell)\in\mathbb{R}^{C_f\times H_\ell\times W_\ell},\quad \ell\in{3,4,5},
\end{equation}
where the per-scale mapping is
\begin{equation}
\label{eq:lmfe_block}
F_\ell=\mathrm{ReLU}!\big(\mathrm{GN}(\mathrm{Conv}{1\times1}(P\ell;\ \text{out}=C_f))\big).
\end{equation}
If a neck path applies stride-2 downsampling, insert an optional depthwise $3{\times}3$ blur before that step.

\textbf{Neck: Dual-branch Neck for Fusion.} To fuse complementary information across scales, we newly designed a dual-module neck with two coordinated blocks. Scale-Sequence Feature Fusion (SSFF) block cross-scale dependencies by treating $\{P_3,P_4,P_5\}$ as an ordered sequence and learning compact 3D kernels with non-negative aggregation. Triple Feature Encoding (TFE) reinforces fine details via large/middle/small receptive-field branches centered on the shallow features. These produce two aligned feature tensors for the head:
\begin{equation}
\begin{aligned}
\label{F_ssff_tfe}
\mathbf{F}_{\mathrm{SSFF}}&=\mathrm{SSFF}(F_3,F_4,F_5),\qquad \\
\mathbf{F}_{\mathrm{TFE}}&=\mathrm{TFE}(F_3,F_4,F_5).
\end{aligned}
\end{equation}
where we define their shapes on the stride-8 grid as
$\mathbf{F}_{\mathrm{SSFF}}\in\mathbb{R}^{C_f\times \frac{H}{8}\times \frac{W}{8}}$, and
$\mathbf{F}_{\mathrm{TFE}}\in\mathbb{R}^{C_f\times \frac{H}{8}\times \frac{W}{8}}$.
The $C_f$ is the fused channel width after alignment. In practice,
SSFF treats $\{P_3,P_4,P_5\}$ as an ordered scale sequence and performs a single
bi-directional fusion pass, while TFE refines high-resolution details by
upsampling deep features to the stride-8 grid and applying channel–spatial
gating on $P_3$. Both tensors are channel-aligned and resolution-matched for
the downstream head.

\textbf{Head: Lightweight Shared‑Convolution Segmentation Head (LMSH).} LMSH employs two shared 3×3 convolutions with GroupNorm to stabilize batch-1 training while minimizing parameters. It supports classification and optional geometric prediction, and generates instance masks either directly or via a prototype–coefficient scheme:
\begin{equation}
\label{eq:LMSH}
\mathbf{p},\ \mathbf{b},\ \mathbf{M}
=\mathrm{LMSH}\!\big(\mathbf{F}_{\mathrm{SSFF}},\ \mathbf{F}_{\mathrm{TFE}},\ P_3,\ P_4,\ P_5\big),
\end{equation}
where $\mathbf{p}$ are class probabilities, $\mathbf{b}$ are optional box/center parameters, and $\mathbf{M}$ are instance masks.


\subsection{C2f Pro Module}
\label{ssec:crf}

\begin{figure}[htbp]
\centering
\subfloat[Traditional C2f module]{
  \includegraphics[width=0.45\textwidth]{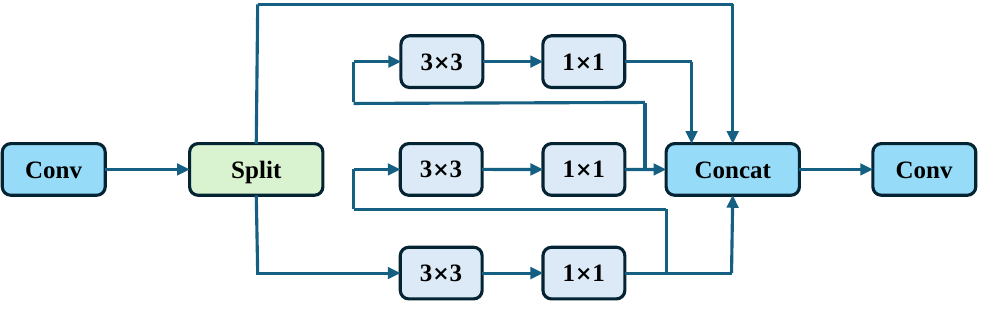}
}
\hfill
\subfloat[Our C2f-Pro module]{
  \includegraphics[width=0.45\textwidth]{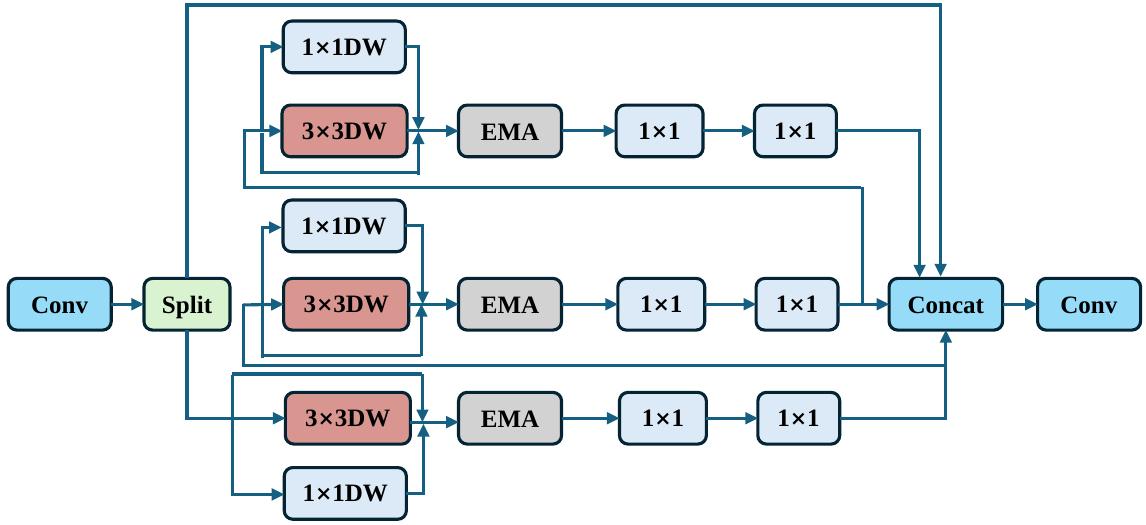}
}
\caption{Comparison between the traditional C2f and our C2f-Pro module.}
\label{fig:c2f_module}
\end{figure}

Spinal endoscopic images exhibit smoothly transitioning textures and thin, low-contrast boundaries between adjacent tissues, which makes purely local aggregation prone to boundary discontinuities or regional adhesion. The original C2f block—built on local convolutions and fixed shortcut interactions—therefore struggles to capture cross-channel and cross-scale dependencies in this setting. To address these limitations under real-time constraints, we redesign C2f into C2f‑Pro by integrating re-parameterized ViT convolutional cells (RVB \cite{wang2024repvit}) and replacing SE \cite{hu2018se} with Efficient Multi‑scale Attention (EMA \cite{shuvo2022efficient}). The details are shown in Figure \ref{fig:c2f_module}. This enhances training-time capacity and global awareness, while structural re-parameterization collapses the multi-branch design into a single path at inference, preserving low latency.

Given an input feature $X\in\mathbb{R}^{C\times H\times W}$ (e.g., $X\in\{P_3,P_4,P_5\}$ after optional channel alignment), C2f‑Pro first compresses channels and forms a shortcut/transform split:
\begin{equation}
\label{eq:c2fpro_split}
X_0=\mathrm{Conv}_{1\times1}(X),\qquad [X_s,\,X_t]=\mathrm{Split}(X_0),
\end{equation}
where $X_0$ is the channel-reduced tensor after the initial $1\times1$ convolution;
$X_s$ is the shortcut branch; and $X_t$ is the transform branch input.

The transform branch passes through three RVB--EMA units in sequence:
\begin{equation}
\label{eq:c2fpro_y}
Y_i=\mathrm{RVB\text{-}EMA}_i(Y_{i-1}),
\end{equation}
where $Y_i$ denotes the output feature of the $i-$th RVB–EMA unit along the transform branch, computed recursively from $Y_{i-1}$ (with $i=1,2,3$) are the successive outputs of the RVB–EMA units on the transform branch with $Y_0=X_t$; these capture progressively enriched representations.


To complement RVB, EMA replaces SE to provide an efficient global context. 
Let $Y_{i-1}\!\in\!\mathbb{R}^{C\times H\times W}$ be the input to the $i$-th unit (with $Y_0=X_t$ from Eq.~\eqref{eq:c2fpro_y}). EMA first computes per-channel global descriptors by average pooling:
\begin{equation}
\label{eq:ema_pool_c2f}
z_c=\frac{1}{HW}\sum_{u=1}^{H}\sum_{v=1}^{W} \big(Y_{i-1}\big)_{c}(u,v),
\end{equation}
where $z\in\mathbb{R}^{C}$ ($c=1,\ldots,C$) summarizes $Y_{i-1}$ over spatial positions $(u,v)$.
A lightweight mixing function $g(\cdot)$ (e.g., a bottleneck MLP or grouped $1\times1$ conv) produces channel weights $a=g(z)\in\mathbb{R}^{C}$, optionally complemented by a spatial gate $S=\sigma(h(Y_{i-1}))\in\mathbb{R}^{1\times H\times W}$ (with $h$ a depthwise $k\times k$ conv and $\sigma$ a sigmoid). The reweighted feature is
\begin{equation}
\label{eq:ema_apply}
\widetilde{Y}_{i-1}(c,u,v)=a_c\, S(u,v)\, Y_{i-1}(c,u,v),
\end{equation}
and the unit output in Eq.~\eqref{eq:c2fpro_y} is then $Y_i=\mathrm{RVB}\big(\widetilde{Y}_{i-1}\big)$.
Thus, EMA is the attention substep inside each RVB--EMA block that transforms $Y_{i-1}$ before the RVB convolutional mixing; it operates on the same tensors defined earlier ($Y_{i-1},Y_i$) and introduces only the auxiliary variables $z$, $a$, and $S$ for channel/spatial gating. In practice, we instantiate EMA at deeper strides ($s\!\in\!\{16,32\}$) to favor the accuracy--latency trade-off.


\subsection{Neck}
\label{ssec:neck}

Spinal endoscopy requires a neck that fuses multi-scale features while preserving thin, low-contrast boundaries under strict real-time constraints. Building on the aligned outputs of Backbone, we structure the neck as two tightly coupled stages: first, Sequence-Style Scale Fusion (SSFF) exchanges information across scales in a single pass; then, Target-Focused Enhancement (TFE) leverages the fused semantics to refine the highest-resolution features where small structures reside. This top-down flow—from global fusion to local emphasis—keeps the design coherent and latency-aware.

\subsubsection{Sequence-Style Scale Fusion (SSFF)}
\label{sssec:ssff}

With inputs standardized by Backbone, SSFF treats the pyramid as a short sequence indexed by scale and performs a single bi-directional pass to communicate both top-down and bottom-up information. 

Let $F_\ell\in\mathbb{R}^{C_f\times H_\ell\times W_\ell}$ for $\ell\in\{3,4,5\}$ denote features at strides $8,16,32$, ordered as $F=[F_3,F_4,F_5]$. Channel width $C_f$ is already aligned by LMFE.
We summarize each scale into a compact token and infer fusion coefficients via a lightweight mixer:
\begin{equation}
\begin{aligned} 
t_\ell&=\big[\mathrm{GAP}(F_\ell)\,;\ \mathrm{SMP}(F_\ell)\big]\in\mathbb{R}^{d},\qquad \\
T&=[t_3,t_4,t_5],
\label{eq:ssff_tokens}
\end{aligned}
\end{equation}
where $\mathrm{GAP}$ is global average pooling, $\mathrm{SMP}$ is $2{\times}2$ strided mean pooling followed by a linear projection, and $[\cdot\,;\cdot]$ denotes concatenation. A causal/anti-causal mixer $\mathcal{M}$ yields per-scale channel weights and gates:
\begin{equation}
[\alpha_3,\alpha_4,\alpha_5],\ [g_3^\uparrow,g_4^\uparrow],\ [g_4^\downarrow,g_5^\downarrow]=\mathcal{M}(T).
\label{eq:ssff_mixer}
\end{equation}

The fused features are computed by a bi-directional update with lightweight resampling:
\begin{equation}
\begin{aligned} 
\widehat{F}_\ell&=\mathrm{Conv}_{1\times1}\!\big(\alpha_\ell\odot F_\ell\big) 
+\underbrace{\mathrm{Up}\!\big(\mathrm{Conv}_{1\times1}(F_{\ell+1})\big)\odot g_\ell^\uparrow}_{\text{top-down}}\\
&+\underbrace{\mathrm{Down}\!\big(\mathrm{Conv}_{1\times1}(F_{\ell-1})\big)\odot g_\ell^\downarrow}_{\text{bottom-up}},
\label{eq:ssff_fuse}
\end{aligned}
\end{equation}
where $\mathrm{Up}$/$\mathrm{Down}$ use nearest-neighbor resampling aligned to $(H_\ell,W_\ell)$; an optional $3{\times}3$ blur precedes $\mathrm{Down}$ to curb aliasing. For boundary protection at the finest scale, an edge-aware gate may be applied after the upsampled path at $\ell{=}3$. The outputs $\widehat{F}_3,\widehat{F}_4,\widehat{F}_5$ provide globally informed yet resolution-faithful features for the subsequent refinement.

\subsubsection{Target-Focused Enhancement (TFE)}
\label{sssec:tfe}

Given the globally fused features from SSFF, TFE concentrates compute where small anatomical structures are best represented: the highest resolution (stride-8). It transfers deep context back to $\widehat{F}_3$ using low-cost channel and spatial emphasis.

We first upsample deeper features to the stride-8 grid and project to form a semantic prior:
\begin{equation}
S=\mathrm{Conv}_{1\times1}\!\big([\mathrm{Up}(\widehat{F}_4),\ \mathrm{Up}(\widehat{F}_5)]\big)\in\mathbb{R}^{C_f\times H_3\times W_3}.
\label{eq:tfe_prior}
\end{equation}
Channel and spatial gates are then computed as
\begin{equation}
\begin{aligned} 
\mathbf{w}&=\sigma\!\big(\mathbf{W}_2\,\delta(\mathbf{W}_1\,\mathrm{GAP}(S))\big)\in(0,1)^{C_f},\qquad \\
M_s&=\sigma\!\big(\mathrm{DW}_{3\times3}(S)+\mathrm{DW}_{5\times5}(S)\big)\in(0,1)^{H_3\times W_3},
\label{eq:tfe_gates}
\end{aligned}
\end{equation}
where $\delta$ and $\sigma$ denote ReLU and Sigmoid, and $\mathrm{DW}$ are depthwise convolutions.

The gates modulate $\widehat{F}_3$ in a residual form,
\begin{equation}
\widetilde{F}_3=(\mathbf{w}\odot \widehat{F}_3)\odot M_s+\widehat{F}_3,
\label{eq:tfe_refine}
\end{equation}
achieving batch-1 stable enhancement that benefits boundary fidelity. To align feature edges with image cues during training, we add a zero-cost-at-inference gradient-consistency loss:
\begin{equation}
\mathcal{L}_{\mathrm{gc}}=\lambda_{\mathrm{gc}}\ \big\|\ \nabla_{\mathrm{Sobel}}(\widetilde{F}_3^{\mathrm{edge}})-\nabla_{\mathrm{Sobel}}(I)\ \big\|_1,
\label{eq:tfe_grad_loss}
\end{equation}
where $\widetilde{F}_3^{\mathrm{edge}}$ is a $1{\times}1$ projection used only for supervision.

The neck outputs $\{\widetilde{F}_3,\widehat{F}_4,\widehat{F}_5\}$, preserving SSFF’s ordering so downstream heads can attach consistently. In multi-scale heads, channels are aligned to $C_h$ by shared $1{\times}1$ projections and depthwise separable $3{\times}3$ blocks per scale. 



\subsection{Lightweight Shared‑Convolution Segmentation Head }
\label{ssec:lmsh}

The Lightweight Shared Convolution Segmentation Head (LMSH) transforms fused neck features into dense masks with minimal latency. 
In particular, LMSH primarily consumes the refined high‑resolution feature $\widetilde{F}_3$, while optionally incorporating cues from $\widehat{F}_4$ and $\widehat{F}_5$. A shared kernel bank applies identical weights across scales to reduce parameters and encourage cross‑scale consistency. Predictions are generated at stride‑4 or stride‑8 (depending on memory constraints), and the logits are then upsampled to the input resolution.
Let $\{\widetilde{F}_3,\widehat{F}_4,\widehat{F}_5\}$ be the outputs of the neck with channel size $C_f$. We first align channels to a head width $C_h$ using shared $1{\times}1$ projections:
\begin{equation}
Z_\ell=\mathrm{Conv}_{1\times1}^{\mathrm{shared}}(F_\ell),\qquad \ell\in\{3,4,5\},
\label{eq:lmsh_proj}
\end{equation}
where the weights of $\mathrm{Conv}_{1\times1}^{\mathrm{shared}}$ are tied across scales.

Each projected feature is refined by $B$ shared depthwise-separable blocks with residual connections:
\begin{equation}
\begin{aligned}
&\Phi^{(b+1)}_\ell=\Phi^{(b)}_\ell+\mathrm{PW}_{1\times1}\!\Big(\delta\!\big(\mathrm{DW}_{3\times3}(\Phi^{(b)}_\ell)\big)\Big),\quad \\
&\Phi^{(0)}_\ell=Z_\ell,\quad b=0,\ldots,B-1,
\label{eq:lmsh_blocks}
\end{aligned}
\end{equation}
where $\mathrm{DW}$ and $\mathrm{PW}$ denote depthwise and pointwise convolutions, and $\delta$ is ReLU. All block parameters are shared across $\ell$.

To obtain a strong stride-8 representation while keeping compute low, we aggregate features to the $H_3{\times}W_3$ grid:
\begin{equation}
\begin{aligned}
U_3&=\Phi_3^{(B)},\quad \\
U_4&=\mathrm{Up}\big(\mathrm{Conv}_{1\times1}(\Phi_4^{(B)})\big),\quad \\
U_5&=\mathrm{Up}\big(\mathrm{Up}\big(\mathrm{Conv}_{1\times1}(\Phi_5^{(B)})\big)\big), \\
\label{eq:lmsh_align}
\end{aligned}
\end{equation}
and fuse them with a lightweight attention gate guided by $U_3$:
\begin{equation}
\begin{aligned}
A&=\sigma\!\big(\mathrm{Conv}_{1\times1}([U_3,U_4,U_5])\big),\qquad \\
G&= A\odot U_3 + (1{-}A)\odot \mathrm{Conv}_{1\times1}([U_4,U_5]),
\label{eq:lmsh_gate}
\end{aligned}
\end{equation}
yielding the head feature $G\in\mathbb{R}^{C_h\times H_3\times W_3}$. This keeps $U_3$ dominant while still injecting deep context.

LMSH produces logits at stride-4 or stride-1 via a minimal decoder:
\begin{equation}
\begin{aligned}
L_8&=\mathrm{Conv}_{1\times1}(G)\in\mathbb{R}^{C_{\mathrm{cls}}\times H_3\times W_3},\qquad \\
L_4&=\mathrm{Conv}_{3\times3}^{\mathrm{DW\mbox{-}PW}}\!\big(\mathrm{Up}(L_8)\big),
\label{eq:lmsh_logits}
\end{aligned}
\end{equation}
where $C_{\mathrm{cls}}$ is the number of classes. If full-resolution masks are required, we apply one or two additional nearest-neighbor upsamplings with a final $1{\times}1$ convolution.

To further enhance thin structures, an auxiliary edge stream supervises high-frequency content without adding runtime cost at inference:
\begin{equation}
\begin{aligned}
&E=\mathrm{Conv}_{1\times1}\!\big(\nabla_{\mathrm{Sobel}}(G)\big),\qquad \\
&\mathcal{L}_{\mathrm{edge}}=\lambda_{\mathrm{edge}}\cdot\mathrm{BCEWithLogits}(E,\ \mathrm{Edge}(Y)),
\label{eq:lmsh_edge}
\end{aligned}
\end{equation}
where $\mathrm{Edge}(Y)$ is a binary edge map derived from the ground truth mask $Y$.

\section{Experiments}
\label{sec:Experimental}
\subsection{Dataset Definition}
\label{ssec:dataset}

To evaluate LMSF-A under realistic intraoperative conditions, we built a spinal endoscopic instance-segmentation dataset from percutaneous endoscopic lumbar discectomy (PELD) cases recorded on an IPS710A endoscopic system at Xiangyang Central Hospital. 
Digital videos from 61 patients were exported and parsed into frames. A standardized sampling-and-filtering pipeline was then applied: every video was sampled at 25 frames per second, followed by removal of frames with motion blur, severe overexposure, dominant instruments, or insufficient anatomical content. After screening, we retained 10 images per patient, yielding a total of 610 images at 1080×720 resolution (approximately 10 frames per case). The dataset can be download on 
\url{https://drive.google.com/file/d/1UXsnQyDAivstFrDTFr5t_YmmQGsx2m3P/view?usp=sharing}.
This study was approved by the Institutional Review Board (IRB) under protocol 2025-199.

\begin{figure}[!htb]
\centering
\includegraphics[width=3.5in]{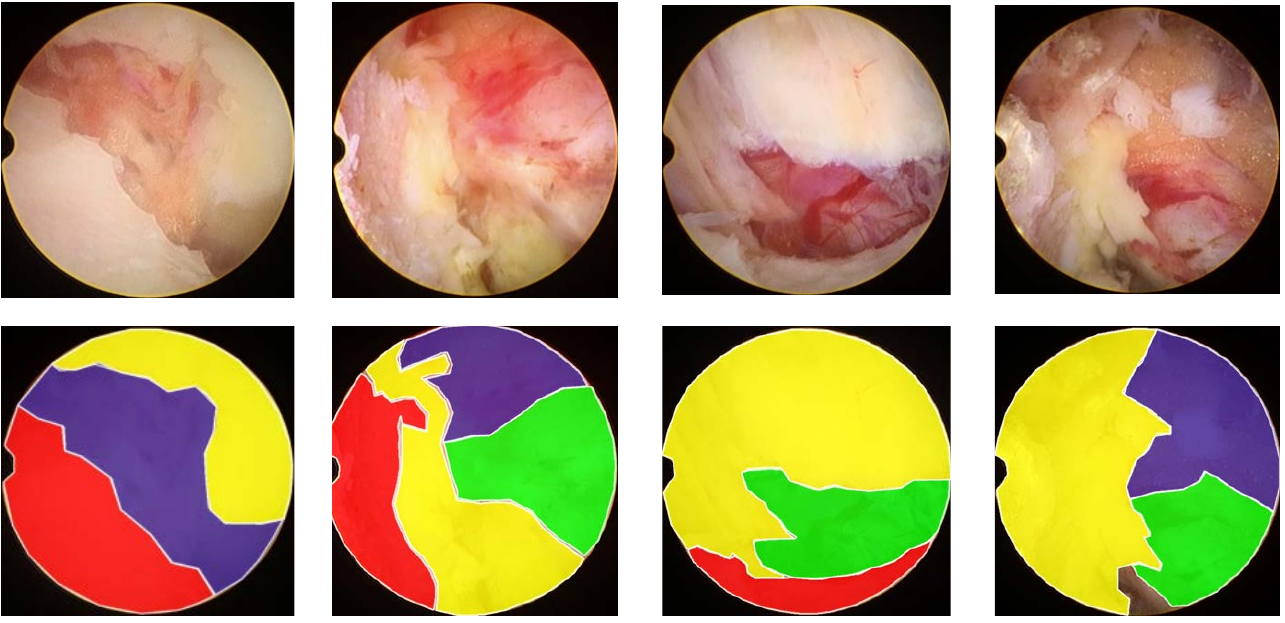}
\caption{Representative samples from the PELD instance‑segmentation dataset. Top row: raw endoscopic frames captured at 1080$\times$720 resolution. Bottom row: corresponding pixel‑wise, instance‑level annotations for four surgical structures—adipose tissue (purple), bone (red), ligamentum flavum (yellow), and nerve (green). Examples illustrate typical variations in viewpoint, illumination, bleeding, and specular highlights encountered during percutaneous endoscopic lumbar discectomy.}
\label{fig:dataset report}
\end{figure}

\textbf{Annotation protocol.}
Curated frames were annotated in LabelMe under the guidance of spine surgeons. Four clinically relevant structures were labeled with pixel-wise, instance-level masks: adipose tissue, bone, ligamentum flavum, and nerve (see Figure~\ref{fig:dataset report}). To control selection bias and enforce consistency, each case was first annotated by a trained rater and then independently reviewed by a second senior surgeon; disagreements were resolved by discussion before data freeze. Colors are used only for visualization (purple: adipose; red: bone; yellow: ligamentum flavum; green: nerve). The training pipeline consumes masks exclusively without relying on color cues.

\begin{figure}[!htb]
  \centering
  \includegraphics[width=0.98\linewidth]{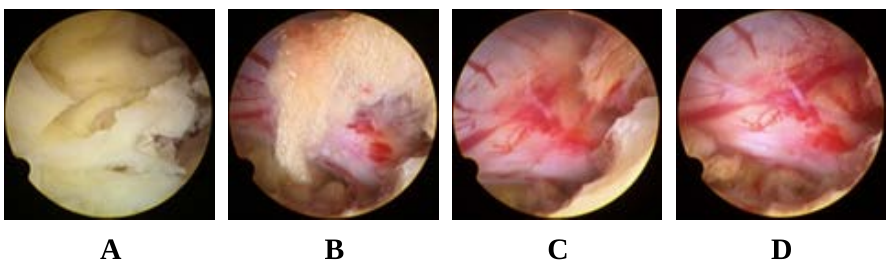}
  \caption{Percutaneous endoscopic lumbar discectomy (PELD) procedural progression. (A) Establishment of the working channel with ligamentum flavum obscuring deeper structures. (B) Partial flavectomy exposing subcutaneous fat and adjacent neural tissue. (C) Further clearance revealing bony landmarks and an intact nerve. (D) Adequate decompression of the compressed nerve root. Typical endoscopic artifacts include narrow field of view, specular highlights, irrigation/blood contamination, and frequent instrument occlusions.}
  \label{fig:procedural}
\end{figure}

\textbf{Procedural Coverage and Visual Challenges.}
These labeled images span the typical PELD workflow and its visual challenges (Figure~\ref{fig:procedural}). Early frames commonly show the establishment of the working canal with the ligamentum flavum covering deeper layers; subsequent steps include partial flavectomy that exposes subcutaneous fat and adjacent neural tissue; further clearance reveals bony landmarks and an intact nerve; and the procedure concludes with adequate decompression of the compressed nerve root. Consequently, the dataset naturally contains narrow fields of view, specular highlights, irrigation/blood contamination, and frequent instrument occlusions—conditions that stress-test small-structure segmentation.

\subsection{Implementation Details}\label{ssec:details}

\textbf{Experimental Settings.}
All experiments were conducted on a single NVIDIA GeForce RTX 3060 GPU with 12\,GB memory. Our software stack consisted of Python~3.10 and PyTorch~2.10, with CUDA~12.1 for GPU acceleration. Input images were resized to $640\times640$ pixels. We trained each model for 300 epochs using a batch size of 8. Stochastic Gradient Descent (SGD) was adopted as the optimizer with momentum set to 0.9 and an initial learning rate of $1.0\times10^{-3}$. Unless otherwise specified, all other hyperparameters followed the defaults of the corresponding implementations.

\textbf{Dataset Split and Statistics.}
For fair and reproducible evaluation across methods, we applied a stratified split that preserves the relative frequency of all four classes. The 610 images are divided into train/val/test with an 8:1:1 ratio (490/60/60). Because a single frame may not include every category, per-class instance counts are lower than the number of images. The final statistics are summarized in Table~\ref{tab:dataset}, and this fixed split is used for all experiments reported in this paper.

\begin{table}[!htb]
\centering
\caption{Dataset split and per-class instance counts. The split is stratified to balance class presence across subsets.}
\label{tab:dataset}
\setlength{\tabcolsep}{12pt}
\begin{threeparttable}
\begin{tabular}{lrrrr}
\midrule
\toprule[1pt]
\# Class  & Train  & Val  & Test  & Total  \\
\hline
Adiposity  & 353 & 40 & 42 & 435 \\
Bone  & 224 & 28 & 25 & 277 \\
Ligament  & 433 & 45 & 32 & 510 \\
Nerve  & 365 & 47 & 43 & 455 \\
\hline
\# Images  & 490 & 60 & 60 & 610 \\
\midrule
\bottomrule[1pt]
\end{tabular}
    \begin{tablenotes}
    \footnotesize
     \item[*]\textbf{Remark}: \#Class and \#Images denote the number of classes and the number of images in each split, respectively.
      \end{tablenotes}
\end{threeparttable}
\end{table}

\begin{figure*}[!htb]
  \centering
  \includegraphics[width=\textwidth]{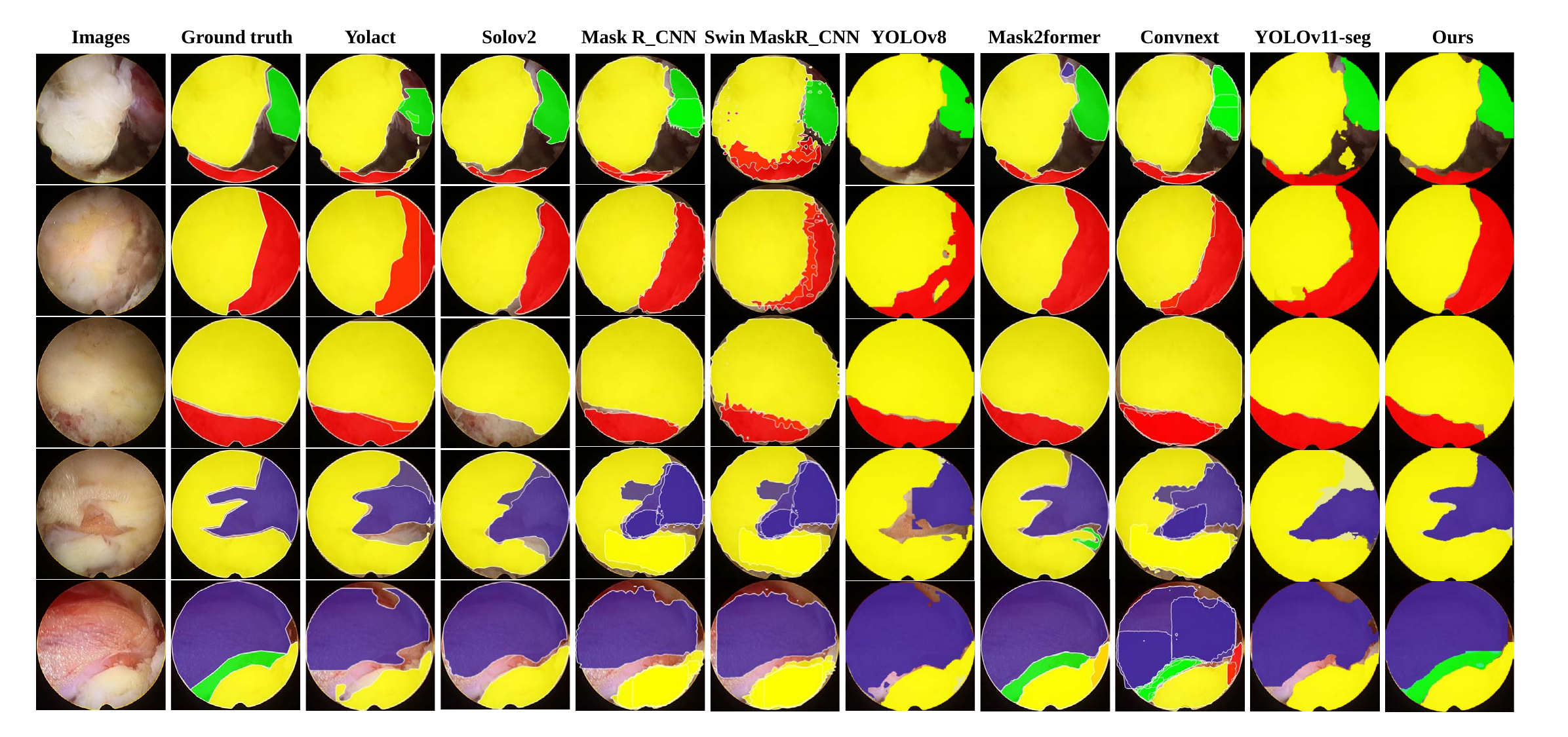}
  \caption{Qualitative comparison on the PELD instance-segmentation dataset.}
  \label{fig:main result}
\end{figure*}

\textbf{Evaluation Metrics}
We evaluate models using mAP, parameters, FLOPs, and FPS. mAP measures the detection/segmentation precision following COCO: $\mathrm{AP}_{50}$ uses IoU$=0.50$, and $\mathrm{AP}_{50\text{--}95}$ averages AP over IoU thresholds 0.50–0.95 (step 0.05). Parameters indicate the size of the model. FLOPs count the operations for one forward pass at the given input. FPS is the inverse of average per-image inference time and, unless noted, excludes data loading and postprocessing. The FPS is measured on the same hardware as in Section~\ref{ssec:details} with batch size 1. Parameter counts and FLOPs are computed in the final trained models at $640\times640$ for fair comparison.


\subsection{Main Results} \label{ssec:Results}

To comprehensively evaluate the proposed method for instance segmentation on spinal endoscopic images. i.e., PELD dataset, we illustrated in Section \ref{ssec:dataset}.
We perform comparative experiments on our PELD  dataset against representative baselines: Mask R-CNN \cite{he2017mask}, SOLOv2 \cite{wang2020solov2}, YOLACT \cite{bolya2019yolact}, and Swin Mask R-CNN \cite{liu2021swin}. These models encompass two-stage and one-stage detectors, as well as Transformer-based architectures, offering robust and diverse references.
﻿
All methods are trained and tested under the same data split and training settings, as reported in Table \ref{tab:dataset}. We report mAP$_{50}$, mAP$_{50-95}$, Parameters, FLOPs, and FPS to assess localization accuracy and model efficiency jointly. The results are summarized in Table \ref{tab:main results}.

\begin{table}[!htbp]
 \centering
 \footnotesize
 \caption{Quantitative comparison of different instance segmentation models on the PELD dataset.}
 \label{tab:main results}
 \begin{threeparttable}
 \resizebox{\linewidth}{!}{
 \begin{tabular}{lcccc}
   \midrule
\toprule[1pt]
    Methods & mAP$_{50}$ ↑ & mAP$_{50-95}$ ↑ & Precision ↑ & A.R. ↑ \\
    \midrule
    Mask R-CNN \cite{he2017mask}        & \underline{75.9} & 41.1 & 74.8 & 61.9 \\
    YOLACT \cite{bolya2019yolact}       & 70.8 & 31.7 & 68.5 & 59.2 \\
    SOLOv2 \cite{wang2020solov2}        & 73.1 & 41.8 & 75.8 & 67.3 \\
    Swin Mask RCNN \cite{liu2021swin}   & 74.4 & 36.5 & 75.2 & 68.4 \\
    Mask2former \cite{cheng2022masked}  & 74.8 & \textbf{43.4} & 78.4 & 65.6 \\
    Convnext \cite{woo2023convnext}     & \textbf{76.0} & 42.4 & 77.8 & \underline{70.6} \\
    YOLOv8-seg \cite{varghese2024yolov8} & 72.0 & 40.9 & 75.4 & 70.3 \\
    YOLOv11-seg \cite{yolo11_ultralytics} & 70.7 & 40.1 & \textbf{80.6} & 61.8 \\
    LMSF-A & 74.4 & \underline{43.0} & \underline{78.9} & \textbf{73.3} \\
\midrule
\bottomrule[1pt]
 \end{tabular}
 }
 \begin{tablenotes}
   \footnotesize
   \item[*]\textbf{Remark}: $\uparrow$($\downarrow$) indicates that the larger (smaller) the value, the better \\
   the performance; the \textbf{Bold} and \underline{underlined} indicate this metric's best and \\ second-best performance, respectively.
 \end{tablenotes}
 \end{threeparttable}
\end{table}

As reported in Table \ref{tab:main results}, our method delivers best or second-best performance across multiple metrics. 
Traditional two-stage methods (e.g., Mask R-CNN \cite{he2017mask}) provide stable accuracy but suffer from slow inference and large parameter counts. One-stage approaches (e.g., SOLOv2 \cite{wang2020solov2} and YOLACT \cite{bolya2019yolact}) offer higher throughput, yet their performance degrades in complex scenes. Swin Mask R-CNN \cite{liu2021swin} achieves strong accuracy but entails substantial computational cost, limiting its suitability for resource-constrained deployment.

To further assess real-world segmentation quality, we compared predictions on representative spinal endoscopic images. As shown in Figure~\ref{fig:main result}, our approach more precisely captures object contours, produces cleaner and more coherent boundaries, and excels at fine-grained structure recognition. Overall, it achieves a favorable balance between accuracy and efficiency, underscoring its comprehensive performance and application potential for spinal endoscopic instance segmentation.



\subsection{Ablation Study}
\label{ssec:ablation_study}

The ablation evaluates the effectiveness and lightweight design of the proposed modules for instance segmentation on PELD dataset by progressively inserting C2f Pro, the improved Neck, and LMSH into a YOLOv8 baseline. Accuracy (mAP$_{50}$, mAP$_{50\text{--}95}$) is summarized in Table~\ref{tab:ablation_lmsfa}, and model complexity is analysised in Sub-section \ref{ssec:eff_ana}.

\begin{table}[!htb]
\centering
\caption{Ablation study of the main components of LMSF-A on PELD dataset. C2f Pro, Neck, and LMSH denote the three modules. Metrics are reported on the validation set.}
\label{tab:ablation_lmsfa}
\resizebox{\linewidth}{!}{
\begin{threeparttable}
\begin{tabular}{ll l l c c}
\midrule
\toprule[1pt]
Backbone & C2f Pro & Neck & LMSH & mAP$_{50}$ (\%) ↑ & mAP$_{50\text{--}95}$ (\%) ↑ \\
\midrule
\checkmark & --         & --         & --         & 72.0 & 40.9 \\
\checkmark & \checkmark & --         & --         & 72.3 & 41.5 \\
\checkmark & --         & \checkmark & --         & 74.2 & 42.8 \\
\checkmark & --         & --         & \checkmark & 72.0 & 41.2 \\
\checkmark & \checkmark & \checkmark & --         & 72.6 & 42.5 \\
\checkmark & \checkmark & --         & \checkmark & 72.5 & 41.5 \\
\checkmark & --         & \checkmark & \checkmark & 72.2 & 42.3 \\
\checkmark & \checkmark & \checkmark & \checkmark & 74.4 & 43.0 \\
\midrule
\bottomrule[1pt]
\end{tabular}
\begin{tablenotes}
   \footnotesize
   \item[*]\textbf{Remark}: $\uparrow$ indicates that the larger the value, the better the performance; 
 \end{tablenotes}
\end{threeparttable}
}
\end{table}

As reported in Table \ref{tab:ablation_lmsfa}, C2f Pro, which emphasizes structural re-parameterization and efficient multi-scale attention, reduces parameters from 3.2M to 2.5M while slightly improving accuracy from 72.0/40.9 to 72.3/41.5, yielding a better accuracy–efficiency trade-off. The improved Neck chiefly enhances discriminative power and cross-scale consistency; although it barely changes parameter count, it delivers clear accuracy gains—especially on cases with ambiguous boundaries or occlusions—reaching 74.2/42.8 when used alone. LMSH leverages shared convolutions and lightweight normalization to keep batch-1 inference stable while further reducing overhead; with LMSH alone, mAP remains around 72.0/41.2, and when combined with other modules, it maintains or improves accuracy without increasing complexity. Integrating all three modules produces the best overall result—74.4 mAP50 and 43.0 mAP50–95—while keeping parameters low, achieving a strong balance between accuracy and deployment efficiency.


\subsection{Efficiency Analysis}
\label{ssec:eff_ana}
\begin{figure}[!htb]
 \centering
 \includegraphics[width=3.5in]{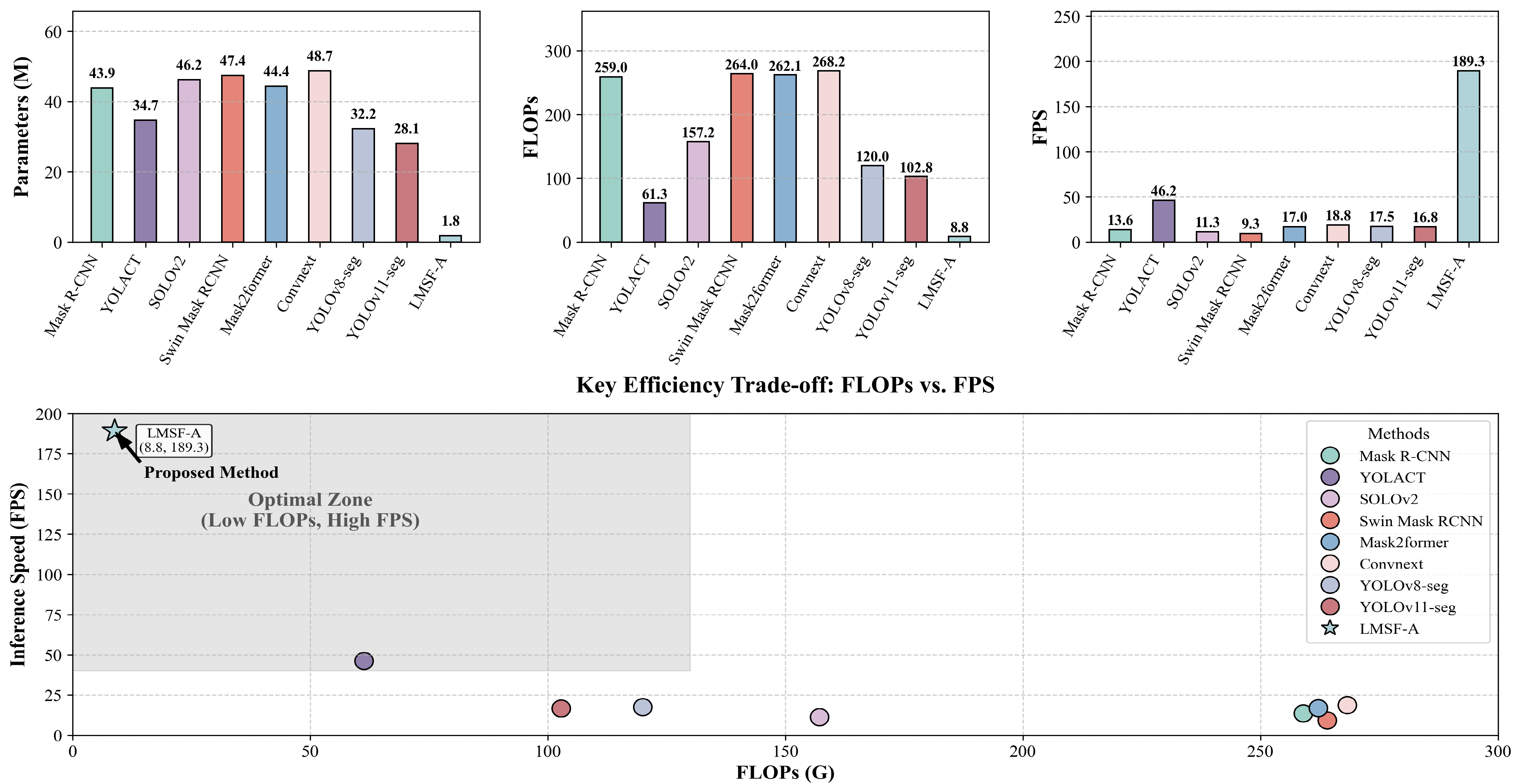}
 \caption{Efficiency comparison across instance-segmentation models on PELD dataset. Top: bar charts of model size (Parameters, M), computation (FLOPs, G), and speed (FPS). Bottom: FLOPs–FPS trade-off scatter plot. LMSF-A lies in the optimal region (low FLOPs, high FPS), achieving 1.8\,M parameters, 8.8\,G FLOPs, and 189.3\,FPS.}
 \label{fig:main_eff}
\end{figure}

Table~\ref{tab:main_eff results} and Figure~\ref{fig:main_eff} compare model size, computation, and speed across representative baselines. LMSF-A attains 1.8\,M parameters, 8.8\,G FLOPs, and 189.3\,FPS, clearly occupying the low-FLOPs/high-FPS region of the trade-off plot. Relative to recent strong detectors (e.g., Mask2Former \cite{cheng2022masked} and ConvNeXt-based method \cite{woo2023convnext} at 44–49\,M parameters and 262–268\,G FLOPs), LMSF-A reduces parameters by over an order of magnitude and cuts computation by roughly 30× while maintaining competitive accuracy.

\begin{table}[!htbp]
 \centering
 \footnotesize
 \caption{Efficiency comparison of instance-segmentation models on PELD dataset.}
 \label{tab:main_eff results}
 \begin{threeparttable}
 \begin{tabular}{lccc}
   \midrule
\toprule[1pt]
    Methods & Parameters (M) ↓ & FLOPs (G) ↓ & FPS ↑ \\
    \midrule
    Mask R-CNN \cite{he2017mask}         & 43.9 & 259.0 & 13.6 \\
    YOLACT \cite{bolya2019yolact}        & 34.7 & 61.3  & \textbf{46.2} \\
    SOLOv2 \cite{wang2020solov2}         & 46.2 & 157.2 & 11.3 \\
    Swin Mask RCNN \cite{liu2021swin}    & 47.4 & 264.0 & 9.3 \\
    Mask2former \cite{cheng2022masked}   & 44.4 & 262.1 & 17.0 \\
    Convnext \cite{woo2023convnext}      & 48.7 & 268.2 & 18.8 \\
    YOLOv8-seg \cite{varghese2024yolov8} & 32.2 & 120.0 & \underline{170.5} \\
    YOLOv11-seg \cite{yolo11_ultralytics}& 28.1 & 102.8 & 168.8 \\
    LMSF-A                                & \textbf{1.8} & \textbf{8.8} & 189.3 \\
\midrule
\bottomrule[1pt]
 \end{tabular}
 \begin{tablenotes}
   \footnotesize
   \item[*]\textbf{Remark}: $\uparrow$ ($\downarrow$) indicates that larger (smaller) values are better; \textbf{Bold} and \underline{underline} denote the best and second-best performance, respectively.
 \end{tablenotes}
 \end{threeparttable}
\end{table}

\begin{figure}[!htb]
 \centering
 \includegraphics[width=3.5in]{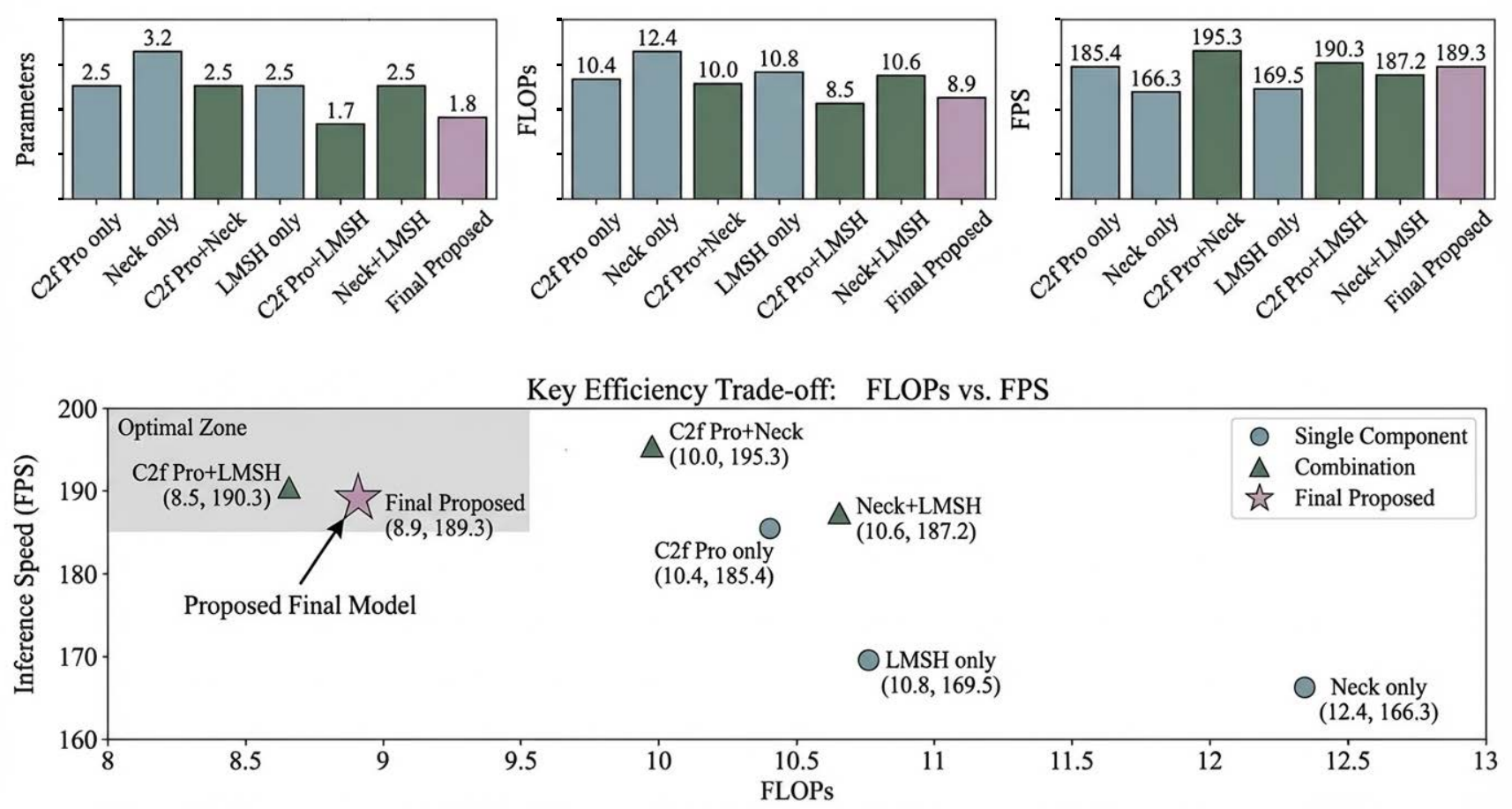}
 \caption{Comparison of model complexity in the ablation study. The bar chart reports Parameters (M) for each configuration, highlighting the reduction achieved by the proposed modules.}
 \label{fig:ablation study analysis}
 \end{figure}

The ablation trend in Table~\ref{tab:ablation_eff} and Figure~\ref{fig:ablation study analysis} highlights the sources of these gains. C2f-Pro and LMSH are the primary contributors to compactness, reducing both parameters and FLOPs without sacrificing speed; their combination yields the tightest efficiency envelope. The improved Neck mainly enhances discriminative capacity: although it adds FLOPs relative to C2f-Pro/LMSH alone, it synergizes with them to sustain high FPS. 

\begin{table}[!htb]
\centering
\caption{Comparison of model complexity in the main components of LMSF-A. C2f Pro, Neck, and LMSH denote the three modules. Metrics are reported on the validation set.}
\label{tab:ablation_eff}
\setlength{\tabcolsep}{8pt}
\resizebox{\linewidth}{!}{
\begin{threeparttable}
\begin{tabular}{ll l l c c c c}
\midrule
\toprule[1pt]
Backbone & C2f Pro & Neck & LMSH & Parameters (M) $\downarrow$ & FLOPs (G) $\downarrow$ & FPS $\uparrow$ \\
\midrule
\checkmark & --         & --         & --         & 3.2 & 12.1 & 170.8 \\
\checkmark & \checkmark & --         & --         & 2.5 & 10.3 & 189.0 \\
\checkmark & --         & \checkmark & --         & 3.2 & 12.4 & 166.3 \\
\checkmark & --         & --         & \checkmark & 2.5 & 10.0 & 195.3 \\
\checkmark & \checkmark & \checkmark & --         & 2.5 & 10.7 & 180.1 \\
\checkmark & \checkmark & --         & \checkmark & 1.9 & 8.4  & 191.1 \\
\checkmark & --         & \checkmark & \checkmark & 2.5 & 10.6 & 187.2 \\
\checkmark & \checkmark & \checkmark & \checkmark & 1.8 & 8.8  & 191.7 \\
\midrule
\bottomrule[1pt]
\end{tabular}
\begin{tablenotes}
\footnotesize
\item[*]\textbf{Remark}: The first row shows the benchmark method (YOLOv8). $\uparrow$ ($\downarrow$) indicates that the larger (smaller) the value, the better the performance. \textbf{Bold} and \underline{underlined} can be used to mark the best and second-best results if desired.
\end{tablenotes}
\end{threeparttable}
}
\end{table}

\subsection{Further Analysis}

\begin{figure}[!htb]
\centering
\includegraphics[width=3.5in]{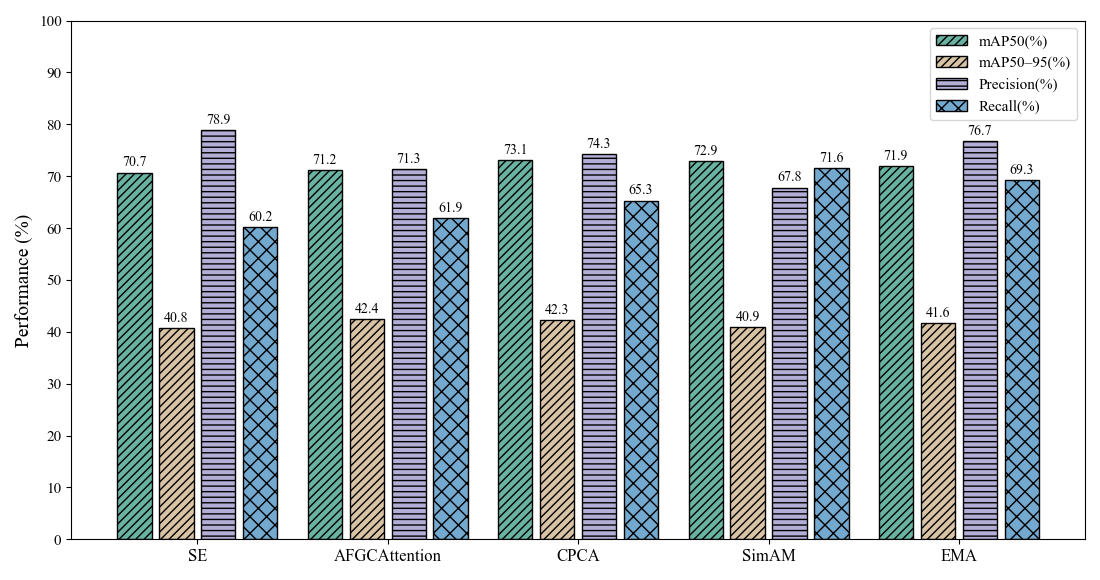}
\caption{Performance of different attention mechanisms for instance segmentation on the PELD dataset.}
\label{fig:per_attent}
\end{figure}

\textbf{Attention mechanism selection.}
To disentangle recognition quality from computational cost, we report accuracy and efficiency separately. Figure~\ref{fig:per_attent} summarizes accuracy (mAP$_{50}$, mAP$_{50\text{--}95}$), Precision, and Average Recall (A.R.) across SE~\cite{hu2018se}, AFGCAttention~\cite{li2025AFGCAttention}, CPCA~\cite{huang2024cpca}, SimAM~\cite{yang2021simam}, and EMA~\cite{ouyang2023efficient}. 
CPCA ~\cite{huang2024cpca} attains the best mAP${50}$ and second-best mAP${50\text{--}95}$ with strong Precision and A.R., while SimAM ~\cite{yang2021simam} yields the highest A.R. SE delivers the highest Precision. 
EMA ~\cite{ouyang2023efficient} offers competitive accuracy (mAP${50}$ and mAP${50\text{--}95}$) with solid Precision/A.R.


\begin{table}[!htb]
\centering
\caption{Complexity and speed of different attention mechanisms.}
\label{tab:attention_complexity}
\begin{threeparttable}
\begin{tabular}{l c c c}
\midrule
\toprule[1pt]
Attention Mechanisms & Parameters (M) ↓ & FLOPs ↓ & FPS ↑ \\
\midrule
SE~\cite{hu2018se}                & 3.2 & 10.2 & 180.2 \\
AFGCAttention~\cite{li2025AFGCAttention} & 3.2 & 12.0 & 183.8 \\
CPCA~\cite{huang2024cpca}            & 3.3 & 12.4 & 182.3 \\
SimAM~\cite{yang2021simam}          & 3.2 & 12.0 & 187.1 \\
EMA~\cite{ouyang2023efficient}              & 2.5 & 10.3 & 189.0 \\
\midrule
\bottomrule[1pt]
\end{tabular}
\begin{tablenotes}
\footnotesize
\item[*]\textbf{Remark}: $\uparrow$ ($\downarrow$) indicates that the larger (smaller) the value, the better the performance; \textbf{bold} and \underline{underlined} mark the best and second-best results per column, respectively.
\end{tablenotes}
\end{threeparttable}
\end{table}


Complementarily, Table~\ref{tab:attention_complexity} shows that EMA has the smallest parameter and FLOP footprint and the highest FPS among peers, making it the most practical choice for real-time endoscopic deployment.

\begin{figure}[!htb]
\centering
\includegraphics[width=3.5in]{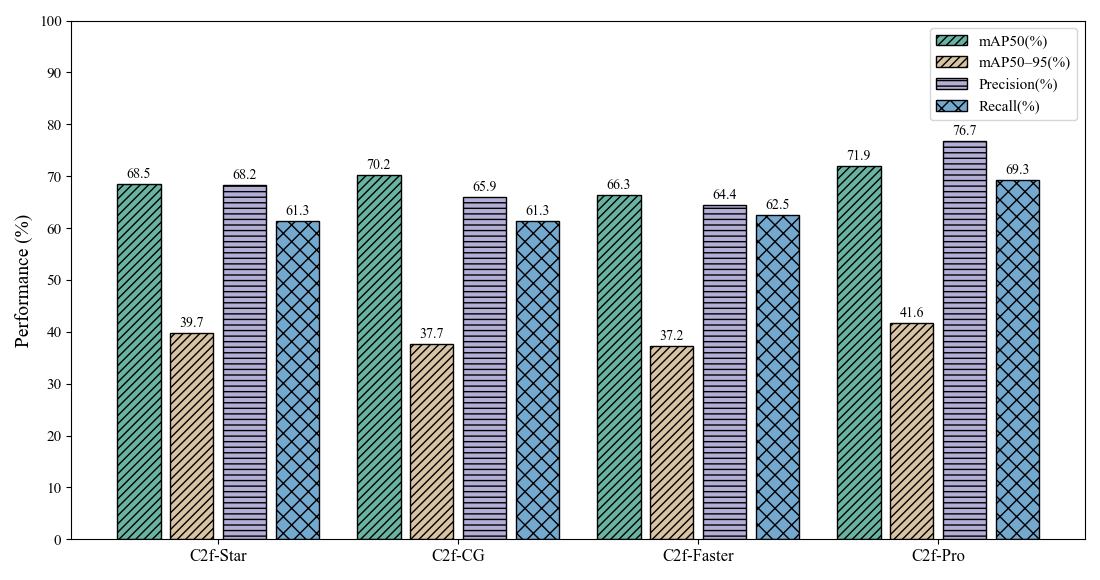}
\caption{Comparative performance of C2f variants in LMSF-A on the PELD dataset. We report mAP$_{50}$, mAP$_{50\text{--}95}$, Precision, and A.R.; higher is better for all metrics.}
\label{fig:performance_c2f}
\end{figure}

\textbf{Backbone variants.}
We evaluate four C2f backbones, i.e., C2f-CG~\cite{wu2020cgnet}, C2f-Faster~\cite{chen2023run}, C2f-Star~\cite{ma2024rewrite}, and our C2f~Pro—in LMSF-A on the PELD dataset. 
As shown in Figure~\ref{fig:performance_c2f}, 
C2f~Pro achieves the best performance on all metrics.
Complementarily, Table~\ref{tab:c2f_modules_eff} summarizes efficiency: C2f-Faster uses the fewest FLOPs and reaches the highest throughput (197.5 FPS), while C2f-CG has the fewest parameters (2.3\,M). 
C2f~Pro offers a favorable trade-off with competitive parameters, near-minimal FLOPs , and strong FPS, aligning accuracy with practical efficiency.


\begin{table}[!htb]
\centering
\caption{Efficiency of different C2f modules. Lower is better for Parameters and FLOPs; higher is better for FPS.}
\label{tab:c2f_modules_eff}
\setlength{\tabcolsep}{10pt}
\begin{threeparttable}
\begin{tabular}{l c c c}
\midrule
\toprule[1pt]
Module & Parameters (M) $\downarrow$ & FLOPs $\downarrow$ & FPS $\uparrow$ \\
\midrule
C2f-CG~\cite{wu2020cgnet}     & \textbf{2.3} & 10.9 & \underline{184.5} \\
C2f-Faster~\cite{chen2023run} & \underline{2.5} & \textbf{10.2} & \textbf{197.5} \\
C2f-Star~\cite{ma2024rewrite} & 2.7 & 10.7 & 182.8 \\
\textbf{C2f~Pro}              & \underline{2.5} & \underline{10.3} & 189.0 \\
\midrule
\bottomrule[1pt]
\end{tabular}
\begin{tablenotes}
\footnotesize
\item[*] $\uparrow$ ($\downarrow$) indicates larger (smaller) is better. \textbf{Bold} and \underline{underlined} mark the best and second-best per column, respectively.
\end{tablenotes}
\end{threeparttable}
\end{table}

Ablations in Subsection~\ref{ssec:ablation_study} further show that replacing the baseline C2f with C2f~Pro reduces parameters and FLOPs while improving mAP$_{50\text{--}95}$, confirming an effective balance between efficiency and representation. 
Implementation-wise, we adopt GroupNorm for batch-1 stability, use stride-2 depthwise-separable convolutions (optionally with blur pooling) for inter-stack downsampling, and expose a standardized channel interface to LMFE and the neck to streamline subsequent fusion.

\begin{figure}[!htb]
\centering
\includegraphics[width=3.5in]{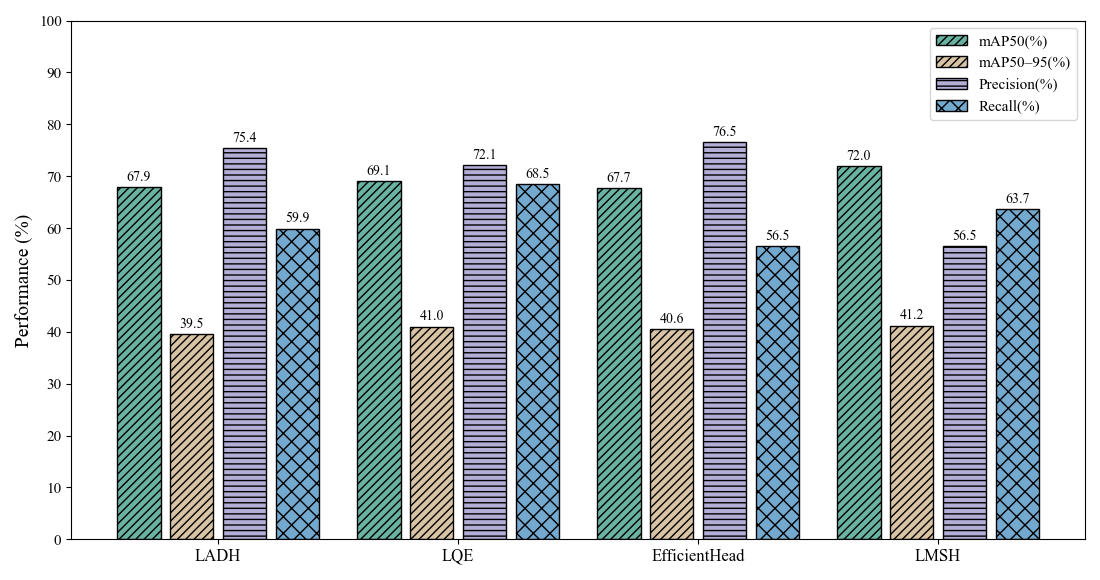}
\caption{Performance of different segmentation heads on PELD instance-segmentation dataset.}
\label{fig:per_seg}
\end{figure}

\textbf{Segmentation head comparison.}
We compare four segmentation heads, e.g., LADH~\cite{LADH}, LQE~\cite{LQE}, EfficientHead~\cite{zhang2023tokenhpe}, and our LMSH on the PELD instance-segmentation benchmark. 
Figure~\ref{fig:per_seg} shows accuracy (mAP$_{50}$, mAP$_{50\text{--}95}$), Precision, and Average Recall (A.R.), while Table~\ref{tab:seg_heads_eff} summarizes efficiency (parameters, FLOPs, and FPS).


Efficiency results in Table~\ref{tab:seg_heads_eff} show that LMSH matches LADH for the fewest parameters (2.5\,M) and achieves near-minimal FLOPs (10.0; second-best) with near-peak throughput (195.3 FPS; second-best). 
Our proposed LADH is marginally lighter in FLOPs (9.9) and slightly faster (196.6 FPS), but its accuracy is notably lower than LMSH on both mAP metrics. 

\begin{table}[!htb]
\centering
\caption{Efficiency of different segmentation heads on PELD dataset.}
\label{tab:seg_heads_eff}
\setlength{\tabcolsep}{10pt}
\begin{threeparttable}
\begin{tabular}{l c c c}
\midrule
\toprule[1pt]
Head &  Parameters (M) $\downarrow$ & FLOPs $\downarrow$ & FPS $\uparrow$ \\
\midrule
LADH~\cite{LADH}           & \textbf{2.5} & \textbf{9.9} & \textbf{196.6} \\
LQE~\cite{LQE}             & \underline{3.2} & 12.0 & 165.3 \\
EfficientHead~\cite{zhang2023tokenhpe} & 4.0 & 11.9 & 172.9 \\
\textbf{LMSH}              & \textbf{2.5} & \underline{10.0} & \underline{195.3} \\
\midrule
\bottomrule[1pt]
\end{tabular}
\begin{tablenotes}
\footnotesize
\item[*]\textbf{Remark}: $\uparrow$ ($\downarrow$) indicates that the larger (smaller) the value, the better the performance; \textbf{bold} and \underline{underlined} mark the best and second-best results per column, respectively.
\end{tablenotes}
\end{threeparttable}
\end{table}

\section{Conclusion}
\label{sec:conclusion}

This work introduces LMSF-A, a lightweight multi-scale attention framework for real-time spinal endoscopic instance segmentation in challenging OR scenes (narrow FoV, glare, fluids, unclear boundaries, large scale changes). It co-designs all stages: i) a C2f-Pro backbone with RepViT-style re-parameterized RVB and efficient multi-scale attention (EMA) for strong features at low cost; ii) a dual neck—SSFF and TFE—for better cross-scale fusion and sharper boundaries; and iii) a Lightweight Multi-task Shared Head (LMSH) with shared convolutions and GroupNorm for stable batch-1 inference and fewer parameters.
We also build a clinician-reviewed PELD instance-segmentation dataset from 61 patients (610 images) with instance masks for adipose tissue, bone, ligamentum flavum, and nerve, enabling standardized evaluation. Experiments show that LMSF-A balances accuracy and efficiency, reaching 74.4 mAP$_{50}$ and 43.0 mAP$_{50\text{--}95}$ on PELD with only 1.8M parameters and 8.8 GFLOPs, running at about 189 FPS (640×640, batch=1). Ablations confirm each component improves accuracy and/or compactness.

\begin{verbatim}


\end{verbatim}


\vspace{11pt}

\vfill

\end{document}